%% file: main.tex
\documentclass{INTERSPEECH2023}

\interspeechcameraready

\usepackage{xcolor}
\usepackage{multicol}
\usepackage{multirow}
\usepackage{soul}
\usepackage{algorithm}
\usepackage{algpseudocode}

\usepackage{comment}

\title{Boosting Punctuation Restoration \\with Data Generation and Reinforcement Learning}
\name{Viet Dac Lai$^{1}$ , Abel Salinas$^{3}$,  Hao Tan$^2$, Trung Bui$^2$, Quan Tran$^2$, Seunghyun Yoon$^2$, \\Hanieh Deilamsalehy$^2$, Franck Dernoncourt$^2$, Thien Huu Nguyen$^1$}

\address{
  $^1$Dept. of Computer Science, University of Oregon, USA\\
  $^2$Adobe Research, USA, 
  $^3$University of Southern California, USA
}
\email{\{vietl,thien\}@cs.uoregon.edu
     asalinas@isi.edu \{haotan,bui,qtran,syoon,deilamsa,franck.dernoncourt\}@adobe.com
     }

\begin{document}

\maketitle

\begin{abstract}
Punctuation restoration is an important task in automatic speech recognition (ASR) which aim to restore the syntactic structure of generated ASR texts to improve readability. While punctuated texts are abundant from written documents, the discrepancy between written punctuated texts and ASR texts limits the usability of written texts in training punctuation restoration systems for ASR texts. This paper proposes a reinforcement learning method to exploit in-topic written texts and recent advances in large pre-trained generative language models to bridge this gap. The experiments show that our method achieves state-of-the-art performance on the ASR test set on two benchmark datasets for punctuation restoration. The source code of this work is publicly accessible at \url{https://github.com/laiviet/pr-rl}.

\end{abstract}

\newcommand{\mytitle}[2]{\multicolumn{#1}{|c|}{\textbf{#2}}}
\newcommand{\mtr}[2]{\multirow{#1}{*}{#2}}
\newcommand{\mtrb}[2]{\multirow{#1}{*}{\textbf{#2}}}
\newcommand{\myrotate}[2]{\parbox[t]{2mm}{\multirow{#1}{*}{\rotatebox[origin=c]{90}{#2}}}}

\input{section1}
\input{section2}
\input{section3}

\input{section4}

\input{section5}

% \clearpage

\bibliographystyle{IEEEtran}
\bibliography{mybib}

% \appendix
% \input{appendix}

\end{document}

%% file: section1.tex
\section{Introduction}

Automatic Speech Recognition (ASR) is a key component in processing audio materials such as audio translation and voice assistant \cite{federmann-lewis-2016-microsoft}, and speech information extraction \cite{cho-etal-2021-streamhover}. Typical ASR systems produce chunks of transcription without any text structures such as sentence and phrase boundaries \cite{jones2003measuring}. As a result, it lowers the readability of the generated ASR texts \cite{jones2003measuring} and severely affects the performance of systems for downstream tasks over this type of text, e.g., information extraction \cite{alam2015comparing}. To address this issue, the Punctuation Restoration (PR) task has been added to the ASR systems \cite{tilk2015lstm} to improve the text readability and the performance of downstream tasks for ASR-generated texts such as question answering \cite{pouran-ben-veyseh-etal-2022-behanceqa}, chitchat detection \cite{lai-etal-2022-behancecc}, and tutorial recommendation \cite{Veyseh2022TutorialRF}. The most recent successful work for PR was all built on top of transformer-based PLMs such as BERT \cite{devlin-etal-2019-bert} and ELECTRA \cite{clark2020electra}.

%together with other post-processing tasks such as true-casing \cite{lita-etal-2003-truecasing}.

% Modern PR systems model the PR task as a word-level sequence labeling task, in which each word is labeled with either a punctuation mark or a NULL label. These models depend on  

% Due to its importance, many studies have been conducted for PR in recent years. 
%Early work employed various combinations of both text-feature and audio features \cite{}. More recent studies mainly develop advanced neural network architectures \cite{}, and integrate external knowledge \cite{}. 
Despite such progress, lacking domain-specific training data is still a major obstacle that hinders the research and development of PR systems for real-world applications \cite{lai-etal-2022-behancepr}. We identify two factors accounting for this issue. First, speech topics involve a unique set of keywords as well as slang in spoken languages. The ASR system and PR system without topic knowledge can be severely affected by the shift of topics in the source audio. Second, unlike other tasks where the unlabeled data is created by humans, the input of PR is generated by an ASR system. This creates a unique dependency that must be addressed by the PR model. Consequently, creating cost-effective datasets for a wide range of domains for PR is highly challenging.

Moreover, naive adoption of available punctuated data is problematic. While large-scale punctuated texts corpora are available, they are mostly written texts (REF texts), which are substantially well-punctuated. In contrast, ASR-generated texts (ASR texts) inherit a substantial amount of noise from both spoken language (e.g., verbal pauses) and the transcription process (e.g., word errors). Accordingly, prior studies have shown that a PR model that was trained on REF texts performed poorly on real-world ASR texts \cite{alam-etal-2020-punctuation}. In other words, directly using readily available written texts does not help to improve the PR model.

To overcome these issues, we introduce a novel data generation method to automatically generate large-scale, high-quality labeled data for PR. In particular, instead of manual annotation, we employ a pre-trained language model, namely GPT2 \cite{radford2019language}, to create synthetic labeled data for PR because generative models like GPT2 can generate punctuated texts that can be converted to labeled data for PR easily. Since the GPT2 model was trained on written texts across diverse topics, this leads to two issues that need to be addressed. 

First, the topics in the generated texts are unconstrained, which is suboptimal for some specific applications, such as gaming livestreaming. As such, we propose a method to control the topic of the generated texts. Instead of unconditional text generation, we feed the GPT2 model with an in-topic seed text, which was sampled from an in-topic unsupervised source. Hence, we encourage the GPT2 model to generate more texts within the initial topic. As a result, we can leverage GPT2's knowledge to obtain unlimited in-topic labeled texts for PR.

Second, the disconnection of the GPT2 model and the target PR model might cause a discrepancy between GPT2-generated texts and the target PR text. Therefore, to improve the quality of the  GPT2-generated data for PR, we propose to further finetune the GPT2 model in parallel with the training of the PR model to generate optimal customized texts for PR. Particularly, we propose a meta-learning framework to consider the GPT2 model as a meta-parameter for the training of the PR model, in which the GPT2 model will be fine-tuned based on the performance of the PR model on the development set. A trivial solution is reinforcement learning, where the reward can be calculated directly from the evaluation metrics of the PR model on the development set, e.g., the F1-score. However, obtaining a reliable, fast reward is challenging due to either the small scale of the evaluation or the computational cost of the evaluation that has to be done at every single iteration. To alleviate this issue, we propose a novel reward function that relies on the gradients of the PR model obtained from the generated texts and the development set. Intuitively, a generated sample should have a higher reward if the PR model's gradients derived from the sample follows the PR expected gradients derived from the development set. Toward this end, in each iteration, after generating synthetic PR data, we compute an average gradient of the PR model over the generated data for each training example. Then, we compute another average gradient of the PR model over a sampled subset of the development set. Finally, the reward for each generated sample is computed using the cosine similarity score between the two gradients. We evaluate the effectiveness of the proposed methods on two benchmark datasets for PR. The experiments show that our model outperforms the strongest baseline on both datasets.

% The GPT2 model is fed with a small chunk of in-topic text to generate more in-topic texts for PR.

% To overcome these issues, instead of annotating more data for some particular domains, we introduce a novel domain-agnostic data generation method to generate labeled PR data automatically. The generated data can inherit a wide range of keywords from the PLM, while the PLM-generated text is also customized to the domain of interest. In particular, we propose to employ a generative PLM named GPT2 to create synthetic data for PR. The GPT2 model is trained on the augmented in-topic data, and then the fine-tuned GPT2 is asked to generate a large amount of labeled semi-in-domain data. This data is then combined with the real in-domain PR data to train a PR model. Moreover, this method allows us to control the seed text that is fed to the GPT2 model, hence controlling the topic in the PLM-generated texts. Hence, this method brings the topic gap between the PLM-generated data and the target PR data.

%  As such, it is necessary that the GPT2 and the PR models can interact with each other so that the PR model can guide the GPT2 model to generate spoken texts instead of written texts. In particular, we proposed a reinforcement learning method to allow the PR model to give feedback to the GPT2 model. So, the GPT2 model can generate texts to optimize the PR model's performance.

%% file: section2.tex
\section{Related Work}

Early PR studies employed syntactic features and prosodic features\cite{szaszak2019leveraging} to train graphical models such as HMM and CRF \cite{tilk2015lstm}. Recent models for PR employed artificial neural networks to model the PR problem as a sequence-to-sequence problem using various network architectures such as convolutional neural network \cite{che-etal-2016-punctuation}, recurrent neural network \cite{tilk2015lstm,kim2019deep}, and transformer \cite{alam-etal-2020-punctuation}. Pretrained language models stand at the core of the recent PR models. There have been variants of pre-trained language models used for PR such as BERT \cite{fu-etal-2021-improving}, RoBERTa \cite{alam-etal-2020-punctuation,courtland-etal-2020-efficient}, ELECTRA \cite{hentschel2021making,chen2021discriminative}, XLM-RoBERTa \cite{chordia-2021-punktuator}, and funnel-transformer \cite{shi2021incorporating}. Recent advance in training and preprocessing leads to many training techniques such as data augmentation \cite{alam-etal-2020-punctuation}, adversarial training \cite{yi2020adversarial}, multitask learning \cite{lin2020joint,hentschel2021making}, self-training \cite{chen2021discriminative}, two-stage training \cite{fu-etal-2021-improving}, and contrastive learning \cite{huang2021token}. External knowledge was also incorporated into the PR model including external punctuated data \cite{fu-etal-2021-improving}, syntactic features \cite{shi2021incorporating} and acoustic features \cite{zhu2022unified}.

%% file: section3.tex
\newcommand{\dunsup}[0]{\mathcal{D}^{unsup}}
\newcommand{\dtrainpr}[0]{\mathcal{D}^{train}}
\newcommand{\ddevpr}[0]{\mathcal{D}^{dev}}
\newcommand{\dtestpr}[0]{\mathcal{D}^{test}}

\newcommand{\bseed}[0]{\mathcal{B}^{seed}}
\newcommand{\bgenpr}[0]{\mathcal{B}^{gen}}
\newcommand{\btrainpr}[0]{\mathcal{B}^{train}}
\newcommand{\bdevpr}[0]{\mathcal{B}^{dev}}
\newcommand{\btestpr}[0]{\mathcal{B}^{test}}

\newcommand{\thetapr}[0]{\theta}
\newcommand{\thetagen}[0]{\omega}

\newcommand{\modelpr}[0]{\mathcal{M}^\theta}
\newcommand{\modelgen}[0]{\mathcal{M}^\omega}

\section{Proposed Approach}

\subsection{Problem Setting}

Similar to prior studies, we model the PR task as a word-level sequence labeling problem. Given a text input sequence $X=\{w_1, w_2, \cdots, w_N\}$ where $N$ is the number of words in the whole sequence, the input $X$ is encoded into vector space using a large language model, parameterized as $f_{\theta}$, as $H=\{h_1, h_2, \cdots, h_N\}$. The ground truth corresponding to the input sequence is $Y=\{y_1,y_2, \cdots, y_N\}$ where $y_i$ belongs to a predefined list of punctuation marks. The model's prediction is formalized as $\hat{Y}=\{\hat{y}_1,\hat{y}_2, \cdots, \hat{y}_N\}$. The PR model, parameterized as $\modelpr$,  is trained using cross-entropy loss: 

$$\mathcal{L}_{CE}=-\frac{1}{N}y_i\text{log}\hat{y}_i$$

We propose a reinforcement learning framework to leverage a generative PLM to adaptively generate PR data for training the PR model. In particular, the learning process involves two models: a PR model $\modelpr$  and a GPT2 model $\modelgen$. It first generates in-topic punctuated text from some seed texts derived from the in-topic unsupervised text. Then, the PR model $\modelpr$ is trained on both the generated data and human-annotated data. Afterward, the GPT2 model $\modelgen$ is finetuned based on the feedback from the PR model to further improve the effectiveness of the generated data. This is achieved by a reinforcement learning algorithm that exploits the agreement between generated data and the development set of the human-annotated data to form the reward function. Algorithm \ref{alg:cap} presents the detail of the proposed method.

\input{table-iwlts}

\subsection{Data Augmentation}
Training/testing data discrepancy is a crucial problem in the punctuation restoration task. The training data that are obtained from written text, however, does not reflect the noise in the actual spoken text that is transcribed by an ASR system. As such, to introduce noise to the text, we augmented the input text using three strategies: \textit{duplication, alternation, deletion}  with an augmentation probability of $\alpha_1, \alpha_2, \alpha_3$ similar to prior work \cite{alam-etal-2020-punctuation}. To fit the very long input sequence into a large language model, the input sequence must be split into shorter segments of the same size. Due to the randomness of the chunking, the predictions of the edge tokens (head and tail of the chunk) might be severely affected due to the lack of preceding or following contexts. To overcome this, we feed additional preceding and following words of a chunk to help the large language model better encodes the sequence for the PR task, especially for predicting the chunk's beginning and ending words. In particular, we concatenate $C$ preceding and $C$ following words to the input sequence, if they are available, resulting in the input sequence $X_C=\{C, X, C\}$ fed to the PR model. We do not predict the labels for these additional tokens to avoid prediction conflict with the preceding and tailing chunks, as well as to prevent the lack of context to recur.
\input{algorithm}

\subsection{Data Generation} 
Due to the limited annotated in-topic data for PR, we proposed a more feasible method to generate an unlimited amount of data for PR using a generative language model, named GPT2. As GPT2 was trained on a massive amount of unsupervised learning text across many topics, it can generate a long piece of text given just a short seed prompt, which controls the topic of the generated text through the seed text given to GPT2. To do that, we obtained the transcripts of the TED-talk from 2013 to 2017 (separated from the IWSLT corpus which covers talks before 2012); then we used this as our unsupervised in-topic corpus for text generation. For the BehancePR corpus, we use the unsupervised text in the development set as the in-topic seed.

In particular, in each iteration, a batch of semi-annotated data $\bgenpr$ is generated by the GPT2 model $\modelgen_{t-1}$ using an in-topic seed $\bseed$. Another batch $\btrainpr$ is sampled from the original PR training data $\dtrainpr$. Finally, the PR model $\modelpr$ is trained on the combined batch of these two batches.

\subsection{Reinforcement Learning} The GPT2 model is helpful in generating well-punctuated in-topic data. However, as the generation is done independently from the PR model, the generated data inherits the written language style from the GPT2 model's memory. As a result, the generated data is not optimal for the PR task, as the ultimate goal of PR is to be used for spoken language. As such, it is necessary for the PR model to give feedback to the GPT2 model so that the GPT2 model can be finetuned in parallel with the training of the PR model. Expectedly, the guidance from the PR model can make the GPT2 model generate more relevant text.

One trivial way to measure the effectiveness of the generated data is the performance of the PR model (e.g., overall F1-score) over the development set. However, as the label in a PR dataset is highly imbalanced, using a discrete measure like F1-score might lead to a high variance reward, hence, inaccurate estimation. Moreover, we aim to train the GPT2 such that the model can learn to generate a sample $\bgenpr$ that resembles the language style in the development set $\ddevpr$. Intuitively, the generated text should be similar to the spoken human language if the gradient updates of the model trained on $\bgenpr$ and $\ddevpr$ are aligned. Formally, the reward $r_i$ for each batch of generated texts $\bgenpr$  is computed as follows: 
\begin{equation}
    r_i = \nabla_{\theta}\mathcal{L}(\bgenpr_i; \theta_{t-1}) \cdot
    \sum_{\mathcal{B}_j \in \ddevpr}\frac{\nabla_{\theta}\mathcal{L}(\mathcal{B}_j; \theta_{t-1}))}{|\ddevpr|}
\end{equation}
where $\mathcal{L}(\mathcal{B}; \theta_{t-1})$ is the cross-entropy of training the PR model $\modelpr_{t-1}$ on the sample $\mathcal{B}$ and $\cdot$ denotes dot product.

Finally, the GPT2 model is trained to maximize negative log-likelihood:
\begin{equation}
\mathcal{L}_G=-\sum_{\mathcal{B}_i \in \bgenpr}r_ilog P(B_i)
\end{equation}

%% file: table-iwlts.tex
\begin{table*}[t]
\centering
% \resizebox{0.9\textwidth}{!}{
\begin{tabular}{|l|l|ccc|ccc|ccc|ccc|}
\hline
    & 
    \mtrb{2}{Model} & 
    \mytitle{3}{Comma} & 
    \mytitle{3}{Period} & 
    \mytitle{3}{Question} & 
    \mytitle{3}{Overall} \\
    \cline{3-14}
    & &P & R & F & P & R & F & P & R & F  & P & R & F  \\
    \hline 
    \hline
    \multirow{10}{*}{\rotatebox{90}{REF}} 
    & ELECTRA-base [110M]  &  69.2 & 76.5 & 72.7 & 89.4 & 90.1 & 89.7 & \textbf{90.7} & 88.6 & \textbf{89.7} &  79.0 & 83.3 & 81.1 \\
    & + Multitask & 76.3 & 76.1 & 76.2 & 88.8 & 89.1 & 89.0 & 88.1 & 84.1 & 86.0 & 82.6 & 82.6 & 82.6 \\
    \cline{2-14}
    % % \cite{alam-etal-2020-punctuation}\\
    & RoBERTa-large [335M] &  76.9 & 75.8 & 76.3 & 86.8 & 90.5 & 88.6 & 72.9 & \textbf{93.5} & 81.9 & 81.6 & 83.3 & 82.4\\
    & + Augmentation &  76.8 & 76.6 & 76.7 & 88.6 & 89.2 & 88.9 & 82.7 & \textbf{93.5} & 87.8 & 82.6 & 83.1 & 82.9 \\ 
    \cline{2-14}
    % % \cite{shi2021incorporating}\\
    & funnel-transformer-xlarge [400M] &  75.5 & 82.4 & 78.8 & 88.7 & 89.0 & 88.9 & 82.4 & 91.3 & 86.6 & 81.7 & 85.8 & 83.7 \\
    & + POS Fusion + SBS &  78.9 & 78.0 & 78.4 & 86.5 & 93.4 & 89.8 & 87.5 & 91.3 & 89.4 & 82.9 & 85.7 & 84.3 \\
    \cline{2-14}
    % % \cite{chen2021discriminative}\\
    & Electra-large [335M] & 76.3 & 81.9 & 79.0 & 89.3 & 90.8 & 90.0 & 79.6 & 93.5 & 86.0 & 82.4 & 86.5 & 84.4 \\
    & + Discriminative Self-Training & 78.0 & \textbf{82.4} & \textbf{80.1} & 89.9 & \textbf{90.8} & \textbf{90.4} & 79.6 & \textbf{93.5} & 86.0 & 83.6 & \textbf{86.7} & \textbf{85.2} \\
    
    \cline{2-14}
    & \textbf{DeBERTa-large} [304M] & 76.2 & 81.4 & 78.7 & 89.5 & 89.8 & 89.6 & 84.0 & 91.3 & 87.5 & 82.6 & 85.7 & 84.1 \\
    & \textbf{+ RL (Ours)} & \textbf{79.3} & 80.8 & \textbf{80.1} & \textbf{90.8} & 90.0 & \textbf{90.4} & 75.9 & 91.1 & 82.8 & \textbf{84.6} & 85.5 & 85.1 \\

    %  \hline \hline
    % BERT + Adversarial &  70.7 & 68.1 & 69.4 & 77.6 & 77.5 & 77.5 & 68.4 & 66.0 & 67.2 & 72.2 & 70.5 & 71.4 \\
    % \hline
    % \cite{hentschel2021making}\\
    \hline
    \hline
    % \myrotate{8}{\textbf{ASR}} 
    \multirow{8}{*}{\rotatebox{90}{ASR}} 
    & ELECTRA-base [110M] & 49.9 & 70.3 & 58.4 & 79.5 & 83.5 & 81.4 & 60.0 & 68.6 & 64.0 & 62.6 & 76.7 & 68.9\\
    & + Multitask & 56.0 & 69.4 & 62.0 & 82.7 & 83.1 & 82.9 & \textbf{69.7} & 65.7 & 67.6 & 68.1 & 76.0 & 71.9\\
    \cline{2-14}
    % \cite{shi2021incorporating} \\
    & funnel-transformer-xlarge [400M]& 52.6 & \textbf{76.5} & 62.3 & 81.2 & 81.8 & 81.5 & 53.1 & 74.3 & 61.9 & 64.1 & 79.1 & 70.8 \\
    & + POS Fusion + SBS & 56.6 & 71.6 & 63.2 & 79.0 & 87.0 & 82.8 & 60.5 & 74.3 & 66.7 & 66.9 & 79.3 & 72.6 \\
    \cline{2-14}
    %  \cite{alam-etal-2020-punctuation}\\
    & RoBERTa-large [355M] & 56.6 & 67.9 & 61.8 & 78.7 & 85.3 & 81.9 & 46.6 & 77.1 & 58.1 & 66.5 & 76.7 & 71.3\\
    & + Augmentation & 64.1 & 68.8 & 66.3 & 81.0 & 83.7 & 82.3 & 55.3 & 74.3 & 63.4
    & 72.0 & 76.2 & 74.0\\ 
    \cline{2-14}
    & \textbf{DeBERTa-large} [304M] & 53.8 & 73.4 & 62.0 & \textbf{83.5} & 81.6 & 82.5 & 60.0 & 79.4 & 68.4 & 66.1 & 77.6 & 71.4 \\
    % + Augmentation & 66.7 & 66.0 & 66.4 & 78.9 & 88.0 & 83.2 & 64.3 & 79.4 & 71.1 & 73.0 & 77.1 & 75.0 \\
    & \textbf{+ RL (Ours)} & \textbf{67.4} & 71.2 & \textbf{69.2} & 82.2 & \textbf{87.3} & \textbf{84.7} & 65.1 & \textbf{82.4} & \textbf{72.7} & \textbf{74.6} & \textbf{79.4} & \textbf{77.0}\\
     \hline
\end{tabular}
% }
\caption{Punctuation prediction performance comparison in terms of precision (P), recall (R), and F1-score (F) on the IWSLT corpus. The Upper half of the table reports the performance on the reference text test set, while the lower half reports the performance of the ASR text test set. Note that ELECTRA-large + Disc Self-Training  model \cite{chen2021discriminative} did not report performance on the ASR text test set.}
\label{table-result-ted}
\end{table*}

%% file: algorithm.tex
\begin{algorithm}[t]
\caption{Reinforcement Learning for PR}\label{alg:cap}
\begin{algorithmic}

\Require $\modelgen, \modelpr, f_{\theta}, f_{\omega}$
\Require $\dunsup$
\Require $\dtrainpr, \ddevpr$

\For{$t$ $<$ max\_iteration}
    
    \State{$\bseed \gets sample(\dunsup)$}
    \State{$\bgenpr \gets \modelgen_{t-1}(\bseed)$} \Comment{Generate data}
    
    \State{$\btrainpr \gets sample(\dtrainpr)$}
    % \State{$\thetapr \gets train(\thetapr, \bgenpr \cup \btrainpr)$} \Comment{Train PR model}
    \State{$\thetapr_t \gets update(\thetapr_{t-1}, \nabla f_{\thetapr}(\bgenpr \cup \btrainpr)$} \Comment{Update PR model}
    
    \State{$\bdevpr \gets sample(\ddevpr)$}
    \State{$grad^{dev} \gets \nabla f_{\thetapr}(\bdevpr)$}
    \State{$grad^{gen} \gets \nabla f_{\thetapr}(\bgenpr)$}
    
    \State{$r=grad^{dev} \times grad^{gen}$} \Comment{Compute reward}

    \State{$\nabla_{\omega}= \sum_{b_i \in \bgenpr}r_i\nabla f_{\thetagen}(b_i)$}
    
    \State{$\thetagen_t \gets update(\thetagen_{t-1}, \nabla_{\omega})$} \Comment{Update GPT2 model}
\EndFor

\end{algorithmic}
\end{algorithm}

%% file: section4.tex
\section{Experiments}

\input{table-behance}

\textbf{Settings}: In this paper, we evaluate our proposed model on two available English datasets that have been used in previous studies. 
\textbf{IWSLT} is the benchmark dataset for the PR task in English. It annotates three prominent punctuation marks: \textit{PERIOD, COMMA, QUESTION}. The IWSLT corpus contains texts derived from TED Talks, which are mainly monologues. The testing set of this corpus contains both reference text (REF), which is well-written text, and transcribed text (ASR) with manually inserted punctuation. Whereas the training set consists of only REF text. The training, development, and test sets contain approximately 2.1M, 300K, and 12K words, respectively. 
\textbf{BehancePR} is a human-annotated dataset for livestreaming videos. It features multiple speakers as well as interaction with a large number of audiences. BehancePR corpus contains only ASR text. The training/development/testing sets contain approximately 1.2M, 34K, and 44K words, respectively. 
The models are evaluated using the standard precision, recall, and F1-score (micro).

\textbf{Hyperparameters}: In this paper, each input word is tokenized using the word-piece tokenizer provided in the PLM. The representation of the first word-piece is collected as the input of the classifier head, which is a fully connected layer, to predict the punctuation.
We employed the DeBERTa-large PLM \cite{he2021deberta} as the encoder of the PR model. The hidden states of the top 8 layers are used as the representation of a token, searched from a pool of \{1,4,8,12\} layers. The GPT2-medium is used to generate the text. The seed texts for the GPT2 model contain 64 consecutive words randomly sampled from these pools. Both models are trained using the Adam optimizer with a learning rate in \{2e-5, 5e-5\}. The augmentation ratios $\alpha_1,\alpha_2,\alpha_3$ are set to 5\%, similar to \cite{alam-etal-2020-punctuation}. We concatenate $C=20$ context words to the head and tail of each chunk. Due to the high cost of evaluating the PR model on the whole development set, in each iteration, we only sample a subset $|B_j|=16$ chunks from $\mathcal{D}_{dev}$ to compute the reward.

\subsection{IWSLT corpus}

\textbf{Baselines}: We compared our model with the state-of-the-art PR models: \textbf{RoBERTa-large+Augmentation} model employs a RoBERTa-large PLM \cite{alam-etal-2020-punctuation}. The input data is augmented using three augmentation strategies: insertion, substitution, and deletion. \textbf{ELECTRA-base+Multitask} \cite{hentschel2021making} is finetuned using additional augmentation detection loss and knowledge distillation loss. \textbf{ELECTRA-large+Discriminative Self-Training} \cite{chen2021discriminative} was self-trained with a discriminator to detect human-annotated data and pseudo-machine-labeled data. \textbf{Funnel-transformer-xlarge+POSFusion} \cite{shi2021incorporating} incorporates additional part-of-speech features from an external neural-network based POS tagger.

\textbf{Results}: Table \ref{table-result-ted} compares the examined models' performance on both the REF test set and the ASR test set. The performance on the REF test set shows us the performance in case the ASR text is close to the written text, while the ASR test shows the actual performance on ASR text.

On the REF test set, ELECTRA-large is the best model among the five examined PLMs with an F1 score of 84.4\%, and it is closely followed by DeBERTa-large (0.3\%  lower). These models leave a large margin to the smaller models such as ELECTRA-base (approx. 3\% lower). Comparing the full models, our DeBERTA-large + RL model gains 1\% over the DeBERTa-large model, achieving 85.1\%. This performance is on par with the ELECTRA-large + Discriminative Self-Training model with a mere margin of 0.1\%.

% On the REF test set, ELECTRA-large is the best model among the five examined PLMs with an F1 score of 84.4\%, and it is closely followed by DeBERTa-large (0.3\%  lower) and funnel-transformer-xlarge (0.7\% lower). This is reasonable given their similar sizes and architectures \cite{clark2020electra,he2021deberta}. These models leave a large margin to the smaller models: RoBERTa-large (approx. 2\% lower) and ELECTRA-base (approx. 3\% lower). Second, comparing the full models, our DeBERTA-large + RL model gains 1\% over the DeBERTa-large model, achieving 85.1\%. This performance is on par with the ELECTRA-large + Discriminative Self-Training model with a mere margin of 0.1\%. Third, comparing the performances for each punctuation, even though the DeBERTa-large + RL loses on QUESTION with substantially lower performance compared to ELECTRA-base and ELECTRA-large models, it yields an identical F1 score to the F1 score of the ELECTRA-large+Discriminative Self-Training model on both COMMA and PERIOD, which account for more than 90\% of the dataset. This experiment shows the effectiveness of the RL training process in providing helpful examples for training the PR model on reference data.

For ASR text, comparing the full models, our DeBERTa-large + RL model (77\% in terms of overall F1) outperforms all the other models at a large margin of 3\% to the highest competitor, RoBERTa-large + Augmentation,  with $p<0.01$. Moreover, without additional training signals or external features, the DeBERTa-large model yields similar performance to other PLMs (e.g., RoBERTa-large and funnel-transformer-xlarge). Furthermore, our proposed model outperforms the other models on all three punctuation marks with a consistently large margin ranging from 1.8\% to 5.1\%, compared to the next highest. These results clearly show the robustness of our proposed RL method to boost the performance of real-world ASR data significantly. The improvement suggests that the RL method has provided helpful training examples to help the model bridge the gap between the REF text and the ASR text in the training and testing data, respectively.

\subsection{BehancePR corpus}

\textbf{Baselines}: We compare our models with the state-of-the-art models that have been evaluated on this corpus. These models include the \textbf{RoBERTa-large} model and its variants with \textbf{Data Augmentation} and \textbf{Conditional Random Field} \cite{alam-etal-2020-punctuation}.

\textbf{Results}: First, we found that data augmentation does not improve the performance of the model trained on the BehancePR dataset. The reason is that the BehancePR dataset's training and testing data are all ASR texts, which is different from the IWSLT corpus in which the training texts are REF texts, and the testing texts are ASR texts. As such, introducing data augmentation skewed the distribution of training and testing data in the BehancePR corpus. Hence, hurting the model's performance. Table \ref{table-result-behance} presents the overall performance of our proposed models on the BehancePR corpus. The DeBERTa-large outperforms the current state-of-the-art  RoBERTa-large+CRF model (62.2\% versus 62.9\%). Furthermore, the DeBERTa-large + RL improves the F1 score from 64.2\% to 65.2\% (+1.0) (statistically significant with $p<0.01$). This again shows the effectiveness of the proposed reinforcement learning methods.

\input{table-ablation}

\subsection{Ablation study}

We perform an ablation study to examine the contribution of each component of the model on the IWSLT ASR test set as shown in Table \ref{table-ablation} (Rows 1-7). Adding the augmentation to the DeBERTa-large model boosts the performance from 71.4\% to 75.0\% (\textbf{+3.6\%}), while \textit{GPT} improves the F1 score from 75.0\% to 75.6\% (\textbf{+0.6\%}). Finally, when we add \textit{RL}, the F1 score jumps from 75.6\% to 77.0\%. These demonstrate that all the proposed components contribute to the improvement. However, data augmentation and RL contribute largely to the performance gain on the IWSLT ASR test set. Finally, to further show the effectiveness of the \textit{RL}, we add it to the RoBERTa-large+Augmentation, resulting in an increase of 1\% in the F1 score. This experiment shows that our RL method is model-agnostic that can be applied to any PR model.

The PR model and the GPT2 model could be finetuned/pre-trained with different strategies. To examine whether finetuned or pre-trained model before the reinforcement learning could further improve the performance of the model. We used the configuration of the full model with GPT2 and RL. However, for the PR model, we trained the PR alone with the same training data for 1 and 2 epochs. Similarly, we pre-trained the GPT2 model on the unsupervised text derived from the training set for the same epochs. Table \ref{table-ablation} (Rows 8-12) reports the performance of these runs. As can be seen from the performance, training/finetuning the model using only PR or GPT2 data significantly hurts the performance of the model. In particular, pretraining a single epoch on PR or GPT2  reduced the performance by 0.4\% to 0.7\%, respectively. Further training the model for one more epoch decreased the performance by 0.4\% to 0.5\%, respectively.

%% file: table-behance.tex
\newcommand{\tablett}[1]{\multicolumn{1}{|c|}{\textbf{#1}}}

\begin{table*}[t]
\centering
\begin{tabular}{|l|ccc|ccc|ccc|ccc|}
\hline
    \mtrb{2}{Model} & 
    \mytitle{3}{Comma} & 
    \mytitle{3}{Period} & 
    \mytitle{3}{Question} & 
    \mytitle{3}{Overall} \\
    \cline{2-13}
    & P & R & F & P & R & F & P & R & F  & P & R & F  \\
    \hline 
    \hline
    % \myrotate{6}{\textbf{ASR}}
    RoBERTA-large \cite{lai-etal-2022-behancepr} & - & - & - & - & - & - & - & - & - &  62.0 & 61.4 & 61.7 \\
    + Augmentation & - & - & - & - & - & - & - & - & - &  63.8 & 60.7 & 62.2 \\
    + CRF & - & - & - & - & - & - & - & - & - &62.2 & 63.5 & 62.9 \\
    + CRF + Augmentation & - & - & - & - & - & - & - & - & - & 61.1 & 62.8 & 62.0 \\
    % ELECTRA-base & 63.3 & 62.4 & 62.9 \\
    \hline
    DeBERTa-large & 61.8 & 58.3 & 60.0 & 65.1 & \textbf{74.6} & \textbf{69.5} & 72.1 & \textbf{56.7} & \textbf{63.5} & 63.7 & 64.8 & 64.2 \\
    % DeBERTa-large + Augmentation &  \\
    \textbf{+ RL (Ours)} &  \textbf{62.1} &\textbf{63.0} & \textbf{62.5} & \textbf{65.9} & 72.4 & 69.0 & \textbf{73.0} & 53.1 & 61.4 & \textbf{64.1} & \textbf{66.2} & \textbf{65.2} \\
    \hline
\end{tabular}
\caption{Performances on the BehancePR test set. Note that \cite{lai-etal-2022-behancepr} did not report the breakdown performance for each type.}
\label{table-result-behance}
\end{table*}

%\newcommand{\tablett}[1]{\multicolumn{1}{|c|}{\textbf{#1}}}

% \begin{table}[t]
% \centering
% % \resizebox{0.4\textwidth}{!}{
% \begin{tabular}{|l|r|r|r|}
%     \hline
%     \tablett{Model} & \tablett{P} & \tablett{R}& \tablett{F1} \\
%     \hline
%     % \myrotate{6}{\textbf{ASR}}
%     RoBERTA-large &  62.0 & 61.4 & 61.7 \\
%     + Augmentation &  63.8 & 60.7 & 62.2 \\
%     + CRF & 62.2 & 63.5 & 62.9 \\
%     + CRF + Augmentation & 61.1 & 62.8 & 62.0 \\
%     % ELECTRA-base & 63.3 & 62.4 & 62.9 \\
%     \hline
%     DeBERTa-large  & 63.7 & 64.8 & 64.2 \\
%     % DeBERTa-large + Augmentation &  \\
%     \textbf{+ RL (Ours)} &  \textbf{64.1} & \textbf{66.2} & \textbf{65.2} \\
%     \hline
% \end{tabular}
% % }
% \caption{Performances on the BehancePR test set. Note that \cite{lai-etal-2022-behancepr} did not report the breakdown performance for each type.}
% \label{table-result-behance}
% \end{table}

%% file: table-ablation.tex
\begin{table}[t]
\centering
% \resizebox{0.83\linewidth}{!}{
\begin{tabular}{|l|l|r|r|r|}
    \hline
    \tablett{Model} & \tablett{P} & \tablett{R}& \tablett{F1} \\
    \hline
    % \myrotate{7}{\textbf{ASR}} 
    RoBERTa-large                   & 66.5 & 76.7 & 71.3  \\
    \hspace{0.2cm}+ Augmentation    & 72.0 & 76.2 & 74.0  \\
    \hspace{0.4cm} + GPT + RL             & 73.3 & 76.7 & 75.0 \\
    \hline
    DeBERTa-large                       & 66.1 & 77.6 & 71.4  \\
    \hspace{0.2cm}+ Augmentation        & 73.0 & 77.1 & 75.0  \\    
    \hspace{0.4cm} + GPT & 74.9 & 76.3 & 75.6 \\
    \hspace{0.6cm} + RL (Full model)  & \textbf{74.6} & \textbf{79.4} & \textbf{77.0} \\
    \hline
    DeBERTa-large + GPT + RL & & & \\
    + PR pretraining (1 epoch)                & 74.5 & 78.5 & 76.4   \\
    + PR pretraining (2 epochs)                & 74.2 & 78.3 & 76.2   \\
    + GPT2 pretraining (1 epoch)               & 75.7 & 77.0 & 76.3   \\
    + GPT2 pretraining (2 epochs)              & 74.5 & 77.2 & 75.8 \\
    \hline
\end{tabular}
% }

\caption{Performances on the IWSLT ASR test set.}
\label{table-ablation}
\end{table}

%% file: section5.tex
\section{Conclusion}

This paper focuses on generating helpful training data for the punctuation restoration task, especially for real-world ASR texts.
We devise a reinforcement learning method to use the GPT2 model to generate additional data to train the punctuation restoration model. This method allows the GPT2 model to learn from real-world ASR text to generate more helpful training examples based on gradient feedback from the PR model. Our model improves PR performance on real-world ASR tests on IWSLT and BehancePR  (+3\% and +2.3\%, respectively). In the future, we would like to extend this research with more advanced gradient feedback to improve the generated data.

\section{Acknowledgement}: This research has been supported by the Army Research Office (ARO) grant W911NF-21-1-0112, the NSF grant CNS-1747798 to the IUCRC Center for Big Learning, and the NSF grant \# 2239570. This research is also supported in part by the Office of the Director of National Intelligence (ODNI), Intelligence Advanced Research Projects Activity (IARPA), via the HIATUS Program contract 2022-22072200003. The views and conclusions contained herein are those of the authors and should not be interpreted as necessarily representing the official policies, either expressed or implied, of ODNI, IARPA, or the U.S. Government. The U.S. Government is authorized to reproduce and distribute reprints for governmental purposes notwithstanding any copyright annotation therein.